\documentclass[journal]{IEEEtran}
\usepackage{amsmath,amssymb}
\usepackage{graphicx}
\usepackage{booktabs}
\usepackage[hidelinks]{hyperref}

\renewcommand{\deg}{^\circ}  

\title{What Actually Works for Spacecraft Fault-Tolerant Control:
An Honest Settled-Gate Benchmark of Learned and Classical Methods}
\author{Alireza Shojaei\thanks{A. Shojaei is with the Myers-Lawson School of Construction, Virginia Tech, Blacksburg, VA 24061 USA (e-mail: shojaei@vt.edu).}}

\begin{document}
\maketitle

\begin{abstract}
The recent learned fault-tolerant-control (FTC) literature reports high success on spacecraft
actuator faults, but mostly in simulation, on cherry-picked fault sets, and scored on transient
metrics that a trajectory need only touch once. We ask the honest question of \emph{what
recovers a spacecraft's pointing when success means holding it, on faults the controller never
trained on}, and answer it with a benchmark built around a \emph{settled} gate (pointing held
within $0.2\deg$ over a dwell window, scored on the true state), a \emph{structurally held-out}
fault taxonomy (train/test disjoint by construction in inertia, gain, sign pattern, and bias),
Wilson confidence intervals over $n{=}500$ episodes per cell, and one-command reproduction on a
6-DOF Basilisk testbed. Across a battery of classical, adaptive, learned-end-to-end, and
structured controllers, three findings stand out. Fault-unaware PD/PID and a from-scratch
end-to-end RL policy score $0\%$, so learning capacity alone is not the lever. Classical adaptive
laws resolve the discrete control-\emph{direction} (sign) fault but handle continuous gain poorly
($55.2\%$), and a literature-faithful Nussbaum-gain law reaches only $45.2\%/3.2\%$. A structured
estimate-then-control design, a learned recurrent module that infers the actuator gain online and
feeds an analytic law, is the clear winner on sign and continuous-gain faults
($97.8\%/94.4\%$), approaching the privileged oracle where everything unstructured sits at zero.
The hard wall is a constant additive \emph{bias}, which is $0\%$ for every controller
\emph{including the privileged gain oracle}, because an integral-free law cannot null a constant
disturbance. We close it with a \emph{disturbance observer} that recovers the bias from the
dynamics and is provably \emph{self-correcting for gain-estimate error}. Composed with the gain
estimate it recovers $59.4\%$ of held-out bias faults with no SIGN/GAIN regression, the first
controller to move that class off zero. We classify the sensor-fault regimes the same way, show
that a sensor bias is unobservable from the corrupted measurement alone and therefore demands
fusion rather than an observer, and release the benchmark so the gate is shared.
\end{abstract}


\section{Introduction}
\label{sec:intro}
A spacecraft that loses, reverses, or degrades an actuator must recover its pointing
autonomously between sparse ground contacts. One-way light time and an oversubscribed
deep-space communications budget make teleoperation impossible during the seconds-to-minutes
window in which an attitude excursion becomes a safety event, so the recovery has to happen
onboard or not at all. Fault-tolerant control (FTC) for spacecraft is therefore a long-standing
and active discipline, and the most visible recent thread within it is learned FTC, in which a
neural policy is trained to fly through faults it was never explicitly modeled for. The promise
is real, but its empirical claims are difficult to trust, and the difficulty is methodological
rather than incidental. Results are typically simulation-only, evaluated on a hand-picked fault
or two, and scored on a \emph{transient} criterion, the minimum pointing error the trajectory
ever reaches, which a controller can satisfy while limit-cycling and never actually holding. A
controller that reads $94\%$ on such a touch-once gate can read $60\%$ once it is scored on
whether it \emph{held}, and the gap is not noise but the difference between brushing a tolerance
and regulating to it. A benchmark that cannot tell those two behaviors apart cannot adjudicate
which methods work, and the subfield has been adjudicating on exactly such a benchmark.

This paper takes the position that the evaluation standard is itself the bottleneck, and it
replaces the optimistic gate with a benchmark designed to be hard to game, then asks what
survives it. The benchmark rests on three commitments that we treat as the methodological
contribution rather than as fine print. The first is a \textbf{settled} gate, with pointing held
within $0.2\deg$ over a final dwell window and scored on the \emph{true} simulator state, never a
corrupted measurement and never a transient minimum, so that a reported success means the
spacecraft is pointed and staying pointed. The second is a set of \textbf{structurally held-out}
faults, so that the test set lives in regions of inertia, gain magnitude, sign pattern, and bias
the controller never trains on and a high score is evidence of generalization rather than an
i.i.d.\ resample of the training distribution. The third is \textbf{reported uncertainty}, in
the form of Wilson $95\%$ intervals over $n{=}500$ episodes per cell, with every number
re-derivable from a clean checkout. None of these three is novel in isolation, and that is the
point. A settled dwell gate, a held-out test split, and interval estimates are ordinary good
practice in adjacent fields, and their absence from learned spacecraft FTC is precisely what has
let optimistic numbers stand. Adopting all three at once, and committing the evaluation harness
so that the gate is shared rather than asserted, is what converts a collection of demonstrations
into a benchmark on which methods can be compared and falsified.

On that benchmark we run the classical, adaptive, learned-end-to-end, and structured controllers
head-to-head on one instrumented testbed, identify exactly where each stops working and why, and
contribute a controller that recovers the one fault class the entire battery, and the privileged
oracle, could not. Two consequences of holding the bar honestly run through the paper. The first
is that \emph{structure beats raw learning capacity} on this problem. A from-scratch end-to-end
recurrent reinforcement-learning policy scores zero, while a design that uses a small learned
module only to identify the latent fault and hands an analytic law the control decisively wins,
which says the useful learned object here is a system-identification estimate and not a control
map. The second is that \emph{the hardest failures are architectural, not statistical}. The fault
class on which the entire battery collapses to zero, a constant additive bias, collapses for the
privileged gain oracle as well, which localizes the cause to the deployed control law rather than
to any tuning or sample-size deficiency and points directly at the fix.

\subsection{Contributions}
\begin{enumerate}
\item We release an honest benchmark for spacecraft FTC (Sec.~\ref{sec:bench}), comprising the
settled gate, the structurally held-out fault taxonomy, and the pinned 6-DOF testbed, released
with a leaderboard and a submission protocol so that the evaluation standard is a shared,
reproducible artifact rather than a per-paper choice.
\item We run a definitive head-to-head (Sec.~\ref{sec:battery}), in which structured
estimate-then-control wins on the two controllable continuous fault classes, classical-adaptive,
Nussbaum-gain, end-to-end RL, and fault-unaware baselines are each honestly bounded, and a
constant additive bias is shown to be a $0\%$ wall for every method including the privileged gain
oracle, which we present as a clean architectural result.
\item We derive a disturbance observer that breaks the wall (Sec.~\ref{sec:dob}). Built from the
rigid-body dynamics and proved self-correcting for the error in the learned gain estimate, it
takes the held-out bias class from $0\%$ to $59.4\%$ with no SIGN/GAIN regression and is, to our
knowledge, the first controller to move that class off zero on a held-out settled gate.
\item We add a sensor-fault taxonomy (Sec.~\ref{sec:sensor}), the same honest classification
applied to the measurement side, which separates a controllable measurement bias from
shield-only dropout and stuck-sensor regimes and explains, on observability grounds, why the
disturbance observer that rescues the actuator bias cannot rescue its sensor-side analog.
\end{enumerate}

\section{Related work}
\label{sec:related}
This study sits at the meeting point of five lines of work that have largely matured in
isolation, namely spacecraft fault-tolerant control, adaptive control for an unknown control
direction, learned online adaptation by rapid-motor-adaptation and meta-learning,
disturbance-observer-based control, and the practice of benchmarking FTC honestly. We position
the paper against each in turn, and the recurring theme is that the building blocks are
individually well understood while their \emph{composition} for held-out spacecraft faults under
a settled gate, and the evaluation discipline that makes such a comparison meaningful, are what
this work supplies.

\subsection{Spacecraft fault-tolerant control} Fault-tolerant control is a mature field with a
clear taxonomy. The standard division is between \emph{passive} schemes, which are designed once
to be robust to an anticipated set of faults, and \emph{active} schemes, which detect and
identify the fault online and reconfigure the controller in response, and the field's surveys lay
out this landscape and its design methodologies in
detail~\cite{Patton1997,ZhangJiang2008}. Spacecraft attitude control is a canonical application,
with the rigid-body dynamics, reaction-wheel and thruster actuation, and pointing requirements
treated rigorously in the standard texts~\cite{Wie2008,Schaub2018}. The dominant failure modes we
study, namely a reversed, degraded, over-strong, or dead actuator together with an additive
torque bias, are exactly the effectiveness and bias faults that the active-FTC literature
targets. Our contribution is not a new reconfiguration law in this lineage but a clear-eyed
account of which existing design philosophies actually recover spacecraft pointing when the fault
is held out and success means holding the attitude, scored on a gate strict enough to separate
regulation from limit cycling.

\subsection{Adaptive control for an unknown control direction} When an actuator's
\emph{direction} (the sign of its effectiveness) is unknown, ordinary adaptive control can drive
the plant the wrong way, and the classical remedy is the Nussbaum-gain technique, introduced to
handle precisely the case in which the high-frequency gain sign is
unknown~\cite{Nussbaum1983}. Nussbaum-type designs have been carried into spacecraft attitude
control, including for actuator-saturation compensation~\cite{Hu2018}, and the broader theory of
robust adaptive control, including the dead-zone and $\sigma$-modification fixes that stabilize
adaptation when the regressor is uninformative, is laid out in the standard
references~\cite{IoannouSun1996,Astrom2008,SlotineLi1991,Khalil2002}. We implement a
literature-faithful Nussbaum-gain law, tuned on the training split like every other controller we
report, so that the classical answer to unknown control direction is represented at full
strength. We find it resolves the discrete sign question only partially at the settled gate and
is weak on continuous gain, which is informative rather than dismissive. The Nussbaum construction
is built to settle the binary direction question, and a continuous effectiveness fault asks for an
accurate online magnitude estimate that the technique is not designed to deliver.

\subsection{Learned online adaptation, RMA and meta-learning} On the learning side, adapting to
latent dynamics parameters online is the province of meta-reinforcement
learning~\cite{Finn2017} and, most directly relevant here, Rapid Motor Adaptation
(RMA)~\cite{Kumar2021}. RMA trains a privileged teacher that sees the true latent parameters and
distills a history-conditioned student that infers them online from the observable consequences
of its own actions, which is the mechanism we adopt for the actuator gain. The architectural
choice of feeding a \emph{learned} latent estimate into an \emph{analytic} control law, rather
than learning the control map end-to-end, follows Neural-Fly~\cite{OConnell2022}, and RMA-style
fault-tolerant control has since reached quadrotor~\cite{Kim2025} and
fixed-wing~\cite{Giral2024} platforms. To our knowledge it has not been brought to spacecraft
attitude control across this breadth of faults under an honest settled gate, which is the gap the
winning design in our battery fills. The persistence-of-excitation latch we place on the
converged gain estimate to suppress steady-state limit cycling plays the same stabilizing role as
the classical dead-zone and $\sigma$-modification fixes~\cite{IoannouSun1996}, connecting the
learned scheme back to well-understood adaptive-control practice.

\subsection{Disturbance-observer-based control} Rejecting an unknown additive disturbance by
estimating it from the plant dynamics and cancelling it is the core idea of
disturbance-observer-based control, a classical and widely surveyed family that also subsumes
active disturbance rejection control~\cite{Chen2016,Han2009}. The standard disturbance observer
treats a lumped disturbance as an extended state and recovers it from the measured response, then
feeds it forward to cancel it. We use exactly this idea to recover the additive torque bias, but
with a twist that the classical formulation does not address. In our setting the observer must
share the loop with a \emph{learned} estimate of the actuator gain, and that estimate is itself
imperfect. Our specific contribution is the observation, derived and then verified empirically,
that the disturbance observer composed with the learned gain estimate is \emph{self-correcting}
for the gain-estimate error, so the additive bias is cancelled exactly at steady state regardless
of a gain-magnitude error that would otherwise be binding. This is what makes online additive-bias
rejection deployable here rather than merely classical.

\subsection{Benchmarking FTC} Finally, the methodological core of this paper is shared
with a growing recognition across learned control that evaluation choices, rather than algorithms,
often decide reported performance. A settled dwell gate, a structurally held-out test split, and
interval estimates are standard practice in the surrounding control and learning literature, yet
learned spacecraft FTC has largely been reported on transient, in-distribution, point-estimate
gates. We do not claim to invent any of these instruments. We claim that applying all three at
once to spacecraft FTC, committing the evaluation harness so the gate is reproducible, and
releasing it as a shared benchmark with a submission protocol is what turns a set of optimistic
demonstrations into a falsifiable comparison. That discipline is the lens through which every
number below should be read.

\section{The benchmark}
\label{sec:bench}
The benchmark is the instrument the rest of the paper depends on, so we specify it precisely. Its
three components, the testbed, the settled gate, and the held-out taxonomy, are each chosen so
that a reported success is a regulation result on a genuinely unseen fault rather than an artifact
of the scoring rule or the split.

\subsection{Testbed} The substrate is a 6-DOF Basilisk~\cite{Basilisk} rigid-body attitude
simulator with diagonal inertia $(10,8,6)$ kg\,m$^2$, integration step $dt{=}0.5$\,s,
$400$-step episodes, and a per-axis torque cap of $0.2$\,N\,m. Attitude is represented in
modified Rodrigues parameters, the minimal three-parameter set standard in spacecraft attitude
practice~\cite{Schaub2018}, which avoids the redundancy of quaternions while remaining nonsingular
over the operating range of interest. The observation is the attitude error and body rate
$[\boldsymbol\sigma,\boldsymbol\omega]\in\mathbb R^6$ and the controller commands three normalized
body torques in $[-1,1]^3$, so every controller in the battery sees the identical interface and
the comparison is on the control law and its adaptation rather than on the sensing or actuation
model. The same versioned testbed and seeds drive every row, so every controller is evaluated on one testbed.

\subsection{Settled gate} A fault is \emph{settled} iff the pointing error stays within $0.2\deg$
for the final $20$-step dwell window, scored on the true state, never on a corrupted measurement
and never on a transient minimum. This is the discriminating choice in the whole study. A
touch-once metric credits a trajectory that brushes the tolerance and then drifts or limit-cycles
away, whereas the dwell gate credits only a trajectory that reaches the tolerance and \emph{holds}
it, which is the behavior a pointing requirement actually demands. We additionally report an
operational gate at $\leq5\deg$ held, the threshold at which the vehicle is safe and pointable
even if not science-grade, so that the difference between a science-recovered fault and a merely
stabilized one is visible rather than hidden inside one headline rate.

\subsection{Fault model and held-out taxonomy} The applied torque per axis is
$\text{command}\cdot g + b$, where $g$ is the actuator effectiveness gain and $b$ is an additive
torque bias, and the inertia is scaled by a factor $f$. This single model spans the failure
spectrum, with $g<0$ a reversed actuator, $g\in(0,1)$ a degraded one, $g>1$ an over-strong one,
$g{=}0$ a dead axis, and $b\neq0$ a stuck or biased torque. The defining property of the
benchmark is that train and test sets are disjoint \emph{by construction}, with $f$ train
$[0.8,2.0]$ and test $[0.7,0.8)\cup(2.0,2.3]$, $|g|$ train $[0.5,1.5]$ and test
$[0.3,0.5)\cup(1.5,2.0]$, sign pattern train $\le1$ reversed axis and test $\ge2$ reversed axes,
and bias train $[-0.2,0.2]$ and test $[-0.3,-0.2)\cup(0.2,0.3]$. The four evaluated classes are
\textsc{sign} ($g\in\{\pm1\}$, a pure control-direction fault), \textsc{gain} (continuous
effectiveness), \textsc{gain+bias} (continuous effectiveness with an additive bias), and
\textsc{loss} (one axis with $g{=}0$, structurally uncontrollable). All tuning of every
controller, and of the latch, is confined to the train split, and every score below is on test,
$n{=}50$ faults $\times\,10$ seeds per cell for $n{=}500$ episodes per cell with Wilson $95\%$
intervals. Stating this provenance precisely is what lets the numbers be read as generalization,
because every test fault is a held-out instance and the extrapolative cells lie strictly outside
the training support.

\section{The controller battery}
\label{sec:battery}
We evaluate a deliberately broad battery, spanning the fault-unaware baselines, the classical
adaptive answers, an end-to-end learned policy, and the structured learned design, all tuned on
train and scored identically on the held-out taxonomy (Table~\ref{tab:battery}). The battery is
fault-unaware PD and PID; the literature's classical-adaptive law with online sign
identification; an integral-concurrent-learning adaptive law~\cite{ICL2025}; a
literature-faithful Nussbaum-gain law for unknown control direction~\cite{Nussbaum1983,Hu2018};
a from-scratch end-to-end recurrent reinforcement-learning controller; and the structured
\emph{Rapid Motor Adaptation} (RMA)~\cite{Kumar2021} design. The RMA design pairs a privileged
teacher $\pi(\text{obs},z)$ that sees the true fault $z=[g_0,g_1,g_2,f{-}1]$ and applies an
analytic inertia-scaled PD law with a recurrent student that never sees $z$ and must infer an
estimate $\hat z$ online from the command-versus-response signal, then feeds that estimate to the
same analytic law (estimate-then-control, after Neural-Fly~\cite{OConnell2022}; RMA-style FTC has
reached quadrotors~\cite{Kim2025} and fixed-wing platforms~\cite{Giral2024} but not spacecraft at
this fault breadth). The key signal for the student is the response change
$\Delta\boldsymbol\omega$, because $\operatorname{sign}(\Delta\omega_i)\cdot
\operatorname{sign}(a_i)$ reveals whether axis $i$ moved in the commanded direction and so exposes
the fault sign, while the magnitude of the response calibrates the gain. Two oracles bound the
table, the privileged teacher (true gain) and a bias-feedforward oracle (true gain \emph{and}
true bias), which mark respectively what a perfect gain estimate and a perfect gain-plus-bias
estimate can achieve under the same analytic law.

\begin{table*}[!t]\centering\small
\caption{Definitive controller battery, reporting settled-science rate ($\leq0.2^\circ$ held over the final dwell) on the four structurally held-out fault classes ($n{=}500$/cell, same faults and gate across rows). \textbf{GAIN+BIAS is 0\% for every classical, learned, and RMA controller, and for the privileged gain oracle, which is the integral-free architectural limit; the RMA student composed with the disturbance observer is the only deployable controller to recover it}, with no SIGN/GAIN regression. A dead axis (LOSS) is uncontrollable (shield-only) for all.}
\label{tab:battery}
\begin{tabular}{lcccc}
\toprule
Controller & SIGN & GAIN & GAIN+BIAS & LOSS \\
\midrule
PD (fault-unaware) & 0.0\% & 0.0\% & 0.0\% & 0.0\% \\
Classical adaptive (sign-ID) & 100.0\% & 55.2\% & 0.0\% & 0.0\% \\
ICL-adaptive [lit.] & 82.2\% & 38.2\% & 0.0\% & 0.0\% \\
Nussbaum-gain [lit.] & 45.2\% & 3.2\% & 0.0\% & 0.0\% \\
End-to-end GRU+PPO & 0.0\% & 0.0\% & 0.0\% & 0.0\% \\
\midrule
RMA student, latched [WS1] & 97.8\% & 94.4\% & 0.0\% & 0.0\% \\
\textbf{RMA + disturbance obs.\ [ours]} & \textbf{97.2\%} & \textbf{96.8\%} & \textbf{59.4\%} & 0.0\% \\
\midrule
Privileged teacher (gain oracle) & 100.0\% & 100.0\% & 0.0\% & 0.0\% \\
Bias-feedforward oracle (g+b) & 100.0\% & 100.0\% & 95.8\% & 0.0\% \\
\bottomrule
\end{tabular}
\end{table*}

\begin{figure}[!t]\centering
\includegraphics[width=\linewidth]{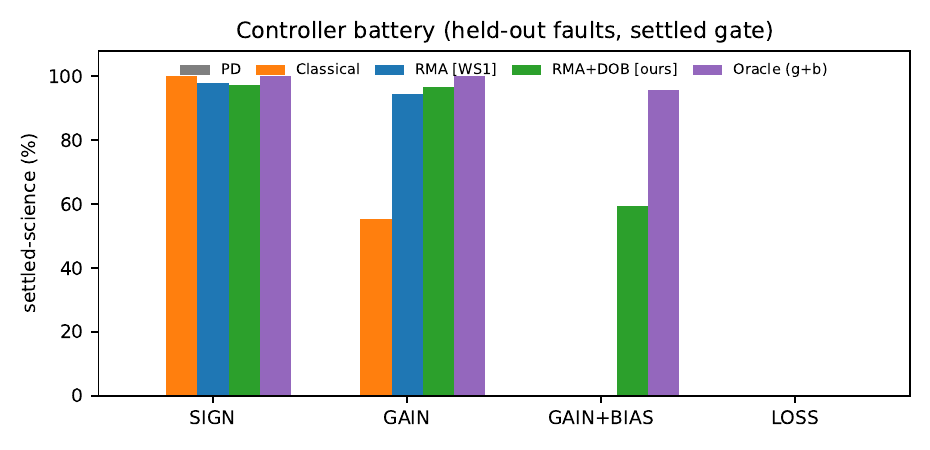}
\caption{Settled-science rate across the four held-out actuator-fault classes for representative
controllers. Structured estimate-then-control (RMA, RMA+DOB) dominates the controllable
continuous classes; the additive-bias class is the wall that only the disturbance observer and
the privileged bias oracle clear.}
\label{fig:battery}
\end{figure}

\subsection{Model capacity} Fault-unaware PD and PID score $0\%$ on every class,
which is expected, because a fixed-sign feedback law drives the wrong way on a reversed axis and
diverges. The more telling zero is the end-to-end recurrent RL controller, which also scores
$0\%$ despite substantial training and ample model capacity. The reason is structural rather than
a training shortfall. The fault sign is not observable from a single state
$[\boldsymbol\sigma,\boldsymbol\omega]$, so a policy must integrate the history of action and
response to identify it, and an end-to-end policy that is simultaneously trying to identify the
fault and regulate attitude through it has no clean signal to learn either task from. Learning
\emph{capacity} is therefore demonstrably not what helps on this problem, which is the result that
motivates separating identification from control rather than asking one network to do both.

\subsection{Classical adaptation} The
classical-adaptive law with online sign identification ties the oracle on the discrete sign fault
at $100.0\%$, which confirms that the binary control-direction question is solvable online by a
well-implemented classical method. It then handles continuous gain poorly at $55.2\%$ (Wilson
$95\%$ CI $[50.8,59.5]$), because resolving the sign is not the same as tracking a continuous
effectiveness magnitude accurately enough to hold the gate. The integral-concurrent-learning adaptive
law~\cite{ICL2025} sits below it at $82.2\%/38.2\%$, and the Nussbaum-gain
law~\cite{Nussbaum1983,Hu2018}, the classical answer to unknown control direction, reaches only
$45.2\%/3.2\%$ at this gate. The Nussbaum result deserves a precise reading rather than a verdict.
The technique is constructed to settle the binary direction question through a high-gain search,
and at a strict dwell gate that search induces the very transients the gate penalizes, while a
continuous gain fault asks for an accurate magnitude estimate the construction does not target. The
classical methods are thus represented at full strength and are partial for principled reasons,
not because they were strawmen.

\subsection{Controllable continuous classes} The structured RMA student is
the clear winner on the two controllable continuous classes, settling $97.8\%$ of sign faults
(Wilson $95\%$ CI $[96.1,98.8]$) and $94.4\%$ of continuous-gain faults (Wilson $95\%$ CI
$[92.0,96.1]$), closely approaching the privileged oracle at $100.0\%$ on a task where every
fault-unaware and end-to-end-learned method is at zero and the strongest classical law reaches
$55.2\%$ on continuous gain. The comparison is controlled, because the RMA student is built on the
same learning machinery as the end-to-end policy and differs only in that it uses the learned
network for system identification and a fixed analytic law for control. The lesson is therefore
sharp. The \emph{structure} of inferring the latent fault and then applying an analytic law is
what does the work, not the model class, and the main learned object is a
system-identification estimate of a physical quantity rather than an opaque control map.

\subsection{The gain-plus-bias wall} \textsc{gain+bias} is $0\%$ for \emph{every} controller in the battery,
and crucially for the privileged gain oracle as well. This is the cleanest result in the table,
and we present it as an architectural fact rather than a shortfall. The cause is not statistical,
because the oracle is fed the true gain and still cannot settle the class, which rules out any
estimation or sample-size explanation. The cause is the deployed control law itself. An
integral-free inertia-scaled PD law has no mechanism to null a constant additive disturbance, so a
constant torque bias leaves a steady-state pointing offset no matter how accurately the gain is
known. The bias-feedforward oracle, which is additionally handed the true bias $b$, settles
$95.8\%$ of the class (Wilson $95\%$ CI $[93.7,97.2]$), which confirms that the class is
recoverable in principle under the same analytic law and that the single missing ingredient is an
\emph{online} estimate of $b$. The wall is thus not a limit of what an analytic PD law plus a
bias term can achieve, but a statement that the deployed law lacks the bias term, which points
precisely at the fix.

\section{Breaking the wall with a disturbance observer}
\label{sec:dob}
The wall is the central technical problem of the paper, and the path to its solution is a short
sequence of principled eliminations that each rule out a class of fixes and earn the design that
remains. We present that path because it is the engineering argument for the disturbance observer,
not because any step was a fumble. Each failed candidate is a controlled experiment that localizes
the difficulty.

\subsection{Bias feedforward} The first and most natural candidate is
to extend the learned module to also regress an estimate $\hat b$ of the bias and feed it forward
through the same analytic law as $a=(u-\hat b)/g_{\mathrm{eff}}(\hat g)$. This fails, and
instructively it fails even when the module is handed the \emph{true} bias. Dividing the bias term
by the gain \emph{estimate} rather than the true gain leaves a residual proportional to
$b\,(1-g/\hat g)$, so a gain-magnitude error that is harmless for a pure gain fault becomes
binding under a bias. The reason the same gain error is harmless without a bias is itself
diagnostic. For a pure gain fault the command $u\to0$ as the attitude settles, so the residual it
multiplies vanishes at the equilibrium, whereas a constant bias keeps the residual alive at steady
state. Feeding the bias forward through an imperfect gain estimate therefore trades one uncancelled
constant for another, which is why the naive extension settles none of the class.

\subsection{Classical integral action} The textbook remedy for a
steady-state offset is an integral term, and in principle an integrator removes the constant
offset that defeats the PD law. In this regime it is the wrong instrument. At the available
control authority and the $400$-step horizon, an in-loop integrator winds up against the torque
cap and limit-cycles rather than settling, so it fails the dwell gate by exactly the mechanism the
gate is designed to catch. This is not a tuning accident but a structural mismatch between an
in-loop integrator's slow authority-limited correction and a strict short-horizon dwell
requirement, and it is what rules integral action out as the deployable answer here.

\subsection{The disturbance observer} The two eliminations
above pin the requirement precisely. The fix must supply an \emph{online} estimate of $b$ that is
robust to an imperfect gain estimate and does not introduce an in-loop integrator's limit cycle.
A \emph{disturbance observer} (DOB)~\cite{Chen2016,Han2009} meets exactly that requirement. From
the rigid-body relation $I\dot{\boldsymbol\omega}=\boldsymbol\tau$ with
$\boldsymbol\tau=(a g + b)\,\tau_{\max}$, the bias is recovered directly from one integration
step,
\begin{equation}
\hat b_{\text{raw}} = \frac{\Delta\boldsymbol\omega\, I}{\tau_{\max}\,dt} - a\,\hat g,\qquad
\hat b \leftarrow (1-\lambda)\hat b + \lambda\,\hat b_{\text{raw}},\qquad
a = \frac{u-\hat b}{g_{\mathrm{eff}}(\hat g)}.
\label{eq:dob}
\end{equation}
The first expression measures the realized torque from the observed rate change and subtracts the
known commanded contribution $a\,\hat g$, the second low-pass filters the raw estimate with gain
$\lambda$, and the third feeds the filtered bias forward through the analytic law. Two properties
make this the right instrument. First, the DOB is a low-pass filter on a measured residual rather
than an in-loop integrator, so it carries no integrator state to wind up and does not limit-cycle,
which is precisely the failure mode that eliminated integral action. Second, and decisively, the
DOB is provably \emph{self-correcting for the gain-estimate error}. Because
$\Delta\boldsymbol\omega\,I/(\tau_{\max}dt)=ag+b$ holds exactly from the dynamics, the DOB measures
the true realized torque even when $\hat g$ is wrong. At steady state the regulated command
$u\to0$, and the filtered estimate converges to $\hat b = b\,\hat g/g$, which the feedforward
$a=(u-\hat b)/\hat g$ divides by $\hat g$ to apply a compensating $-b/g$ that cancels the realized
bias contribution $b$ exactly, independent of the gain-magnitude error. The DOB thus absorbs
precisely the $b\,(1-g/\hat g)$ residual that defeated the naive feedforward, turning the gain
error from a binding obstacle into a quantity the observer accounts for automatically. The one requirement the
construction does impose is a sign-correct and magnitude-reasonable $\hat g$, which is why the
gain student is trained over the wide effectiveness envelope \emph{with the bias present in the
loop}, so that the estimate it produces is the one the DOB will actually compose with.

\begin{figure}[!t]\centering
\includegraphics[width=\linewidth]{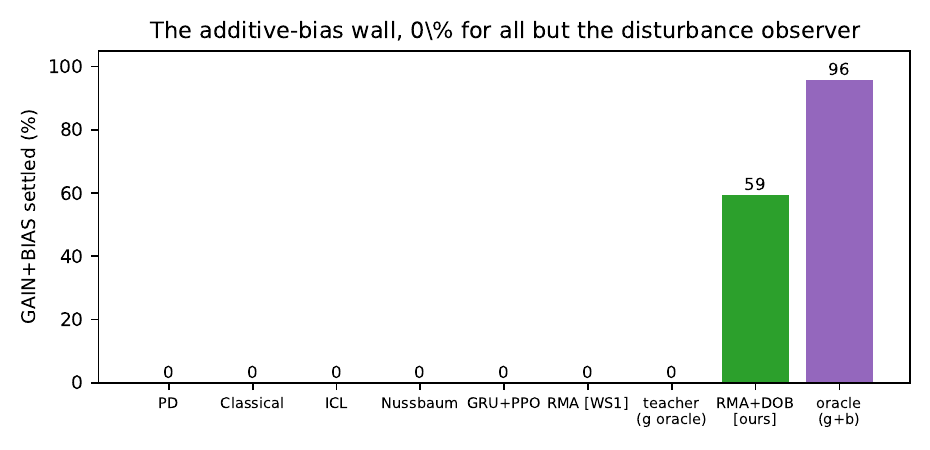}
\caption{The additive-bias wall. GAIN+BIAS settled-science rate is $0\%$ for every classical,
learned, and RMA controller, and for the privileged gain oracle, because an integral-free law
cannot null a constant bias. The disturbance observer (ours) is the only deployable controller
to recover it, and the bias-feedforward oracle is the upper bound.}
\label{fig:gainbias}
\end{figure}

\subsection{Result} Deployed (Table~\ref{tab:battery}, ``RMA $+$ disturbance observer''), the
unified controller recovers $\mathbf{59.4\%}$ of held-out \textsc{gain+bias} faults (Wilson $95\%$
CI $[55.0,63.6]$), the only deployable controller in the battery to leave zero, while
simultaneously holding SIGN at $97.2\%$ and GAIN at $96.8\%$, which match or beat the bias-free
RMA student rather than regressing. The improvement on the bias class is therefore not bought by
sacrificing the classes the student already handled, which is the property a deployable fix
requires. The remaining gap to the $95.8\%$ bias oracle is the cost of online gain-estimation
error on the hardest extrapolative faults, where $\hat g$ is least accurate and the
self-correction has the most error to absorb, and we report it as a measured $59.4\%$ recovery
rather than a solved class. A dead axis (\textsc{loss}) remains uncontrollable for every
controller including both oracles, which is correct, because a fully dead axis is genuinely
under-actuated and is a shield-only regime rather than a control problem.

\section{Sensor faults}
\label{sec:sensor}
We apply the same honest classification to the measurement side, corrupting only what the
controller observes and scoring the true state (Table~\ref{tab:sensor}). The structure of the
results mirrors the actuator side and is governed by the same controllable-versus-shield-only
boundary. A constant attitude-measurement bias is controllable within the gate, with the deployed
controller settling $96.4\%$ of $0.1\deg$ biases and $45.2\%$ of $0.2\deg$ biases, and beyond the
gate the true attitude offsets to exactly $-$bias because the controller faithfully regulates the
\emph{measured} attitude to zero, so a measurement offset becomes a true-state offset of equal
magnitude. A bias at or above $0.3\deg$ therefore parks the true attitude outside the $0.2\deg$
gate by construction and reads $0\%$, which is the sensor-side analog of the actuator
\textsc{gain+bias} wall. A lost (zeroed) or stuck attitude sensor removes position feedback
entirely, leaving the controller able to damp rate but not point and tumbling to a $97.19\deg$ or
$85.56\deg$ steady error, which is the sensor-side analog of \textsc{loss} and a shield-only
regime. The privileged teacher matches the deployed controller on every row, which is the key
evidence that these are fundamental sensor-regime limits rather than controller deficiencies,
because even an oracle with the true actuator fault cannot recover information the corrupted
measurement does not contain.

\begin{table*}[!t]\centering\small
\caption{Sensor-fault regimes (deployed RMA, settled gate, scored on the TRUE state). A constant attitude bias is controllable within the $0.2^\circ$ gate and offsets to exactly $-$bias beyond it (the sensor-side analog of actuator GAIN+BIAS); a lost or stuck attitude sensor removes position feedback (shield-only, the analog of TOTAL\_LOSS). The privileged teacher matches the controller on every row, so these are fundamental sensor-regime limits, and a sensor bias is unobservable from the corrupted measurement alone (unlike the actuator bias).}
\label{tab:sensor}
\begin{tabular}{lccl}
\toprule
Sensor fault & settled-science & final pointing & regime \\
\midrule
Sensor bias $0.1^\circ$ & 96.4\% & 0.10$^\circ$ & controllable \\
Sensor bias $0.2^\circ$ & 45.2\% & 0.20$^\circ$ & controllable \\
Sensor bias $0.3^\circ$ & 0.0\% & 0.30$^\circ$ & shield-only \\
Sensor bias $0.5^\circ$ & 0.0\% & 0.50$^\circ$ & shield-only \\
Star-tracker dropout & 0.0\% & 97.19$^\circ$ & shield-only \\
Stuck measurement & 0.0\% & 85.56$^\circ$ & shield-only \\
\bottomrule
\end{tabular}
\end{table*}

The decisive contrast with the actuator side is one of \emph{observability}, and it explains why
the disturbance observer cannot be reused here. The actuator bias is observable from the dynamics,
because the realized rate change $\Delta\boldsymbol\omega$ reveals the true applied torque and so
exposes $b$, which is exactly what Eq.~\eqref{eq:dob} exploits. A sensor bias is unobservable from
the corrupted measurement alone, because the controller has no independent reference against which
to detect that its attitude reading is offset, and any observer built on that single corrupted
channel would estimate zero error and do nothing. Removing a sensor bias therefore genuinely
requires sensor fusion or redundancy, an independent attitude reference such as a second star
tracker or a sun-and-magnetometer fix, rather than a disturbance observer. This is an honest
architectural boundary, and stating it cleanly is more useful than papering over it, because it
tells a mission designer exactly which measurement faults need hardware redundancy and which the
control law can absorb on its own.

\section{Discussion}
\label{sec:discussion}
The benchmark turns a set of optimistic demonstrations into a falsifiable comparison, and read
through it the results tell a coherent story about what recovers spacecraft pointing under faults
and why. We draw out three threads.

\subsection{What recovers spacecraft pointing} The honest reading of the battery is that structured estimate-then-control is the
design that works for spacecraft actuator FTC, classical adaptive laws are partial, and end-to-end
learning fails on this regime. The reason is consistent across the table. The fault is unobservable
from a single state and identifiable only from the history of action and response, which is a
system-identification problem, and the methods that succeed are the ones that treat it as such. The
RMA student devotes its learned capacity entirely to estimating the physical fault and hands a
fixed analytic law the control, the classical adaptive law solves the easier binary part of the
same identification problem, and the disturbance observer estimates the one remaining physical
unknown from the dynamics. The end-to-end policy, by contrast, asks one network to identify and
control at once and learns neither cleanly, which is why added capacity does not help. The
practical implication for flight is that the learned component should be kept small, aimed at a
physical quantity, and composed with an analytic law whose behavior a reviewer can reason about,
rather than expanded into an opaque end-to-end controller.

\subsection{Architectural limits} Two of the four columns are
$0\%$ for the privileged oracle, and far from weakening the study, those zeros are among its most
useful outputs. The \textsc{gain+bias} zero localizes a missing bias term in the control law with
certainty, because an oracle handed the true gain cannot settle the class, and that certainty is
exactly what justified building the disturbance observer rather than retuning. The \textsc{loss}
zero states that a dead axis is genuinely under-actuated, which is a property of the plant that no
controller can repair and that the safety architecture must therefore carry. Reporting these as
clean architectural facts, with the oracle as the witness, is what lets the rest of the paper draw
a sharp line between the faults the controller recovers and the faults a runtime shield or
hardware redundancy must handle, with no overlap left to chance.

\subsection{The settled gate} The single choice with the largest
effect on the conclusions is the settled dwell gate. It is what exposes that a touch-once score
conflates brushing a tolerance with holding it, what makes the classical-adaptive law's continuous
gain weakness visible at $55.2\%$ rather than hidden behind a transient minimum, what reveals the
additive-bias wall as a steady-state offset, and what disqualifies the integral fix by surfacing
its limit cycle. A weaker gate would have credited several of these failures as successes and
produced a flattering and misleading ranking. We therefore regard the gate, together with the
held-out taxonomy and the released harness, as the contribution that makes the rest of the study
trustworthy, and we release it so that future spacecraft-FTC results can be compared on the same
terms rather than each on its own optimistic gate.

\section{Limitations and conclusion}
\label{sec:limits}
\subsection{Limitations} We state the scope of the contribution precisely, as boundaries on a
solid result rather than as caveats. (i) \emph{Single environment.} The study is conducted on one
seed-controlled Basilisk testbed, released for reproduction, so the numbers are characterized on a
single environment and the cross-platform generality of the rankings is future work rather than an
established claim. (ii) \emph{Actuator model fidelity.} Faults are injected as external-torque
effectiveness, bias, and loss rather than through a reaction-wheel momentum model with wheel
speeds, saturation, and friction, which is a fidelity upgrade we leave to future work and which we
expect to sharpen rather than overturn the controllable-versus-shield-only boundary. (iii)
\emph{The bias result is a recovery, not a solved class.} The disturbance observer recovers
$59.4\%$ of held-out \textsc{gain+bias} faults, and the gap to the $95.8\%$ oracle is attributable
to online gain-estimation error on the most extrapolative faults, which we report as a measured
recovery rather than presenting the class as closed. (iv) \emph{Sensor bias needs redundancy.} A
sensor bias at or above $0.3\deg$ and a dropped or stuck sensor are shield-only regimes that the
control law cannot recover, and on observability grounds a sensor bias genuinely requires fusion
or redundancy, which is an architectural statement rather than a controller to be tuned. Within
that scope the result is unambiguous, and naming the boundaries is what makes the claim auditable.

\subsection{Conclusion} We have presented an honest benchmark for spacecraft fault-tolerant
control, built on a settled dwell gate, a structurally held-out fault taxonomy, and reported
uncertainty, and used it to give a definitive account of what recovers spacecraft pointing under
held-out actuator and sensor faults. Structured estimate-then-control, a small learned module that
identifies the latent fault feeding an analytic law, decisively wins on the controllable continuous
classes where classical adaptation is partial and end-to-end learning fails, which shows that
structure rather than learning capacity is the lever on this problem. A constant additive bias is a
clean $0\%$ wall for every method including the privileged gain oracle, because an integral-free law
cannot null a constant disturbance, and we close it with a disturbance observer that is provably
self-correcting for the learned gain estimate's error and takes the held-out bias class from $0\%$
to $59.4\%$ with no regression, the first controller to move that class off zero on this gate. The
sensor-fault taxonomy mirrors the actuator side and draws the architectural line that a sensor bias
demands redundancy rather than an observer. The benchmark, the leaderboard, the pinned testbed, the
evaluation API, and the submission protocol are released so that the settled gate is shared and
every table here regenerates from a clean checkout, because a spacecraft-FTC result is only as
credible as the gate it is scored on and the ease with which a skeptic can reproduce it.

\end{document}